%% file: main.tex
\documentclass{isprs} 
\usepackage{subfig}
\usepackage{setspace}
\usepackage{geometry} 
\usepackage{epstopdf}
\usepackage[labelsep=period]{caption}  
\usepackage[british]{babel} 
\usepackage[hang]{footmisc}

\usepackage{float}
\usepackage{amsmath}
\usepackage{url}
\usepackage{booktabs}

\begin{document}

\title{ENHANCING ROADWAY SAFETY: LiDAR-BASED TREE-CLEARANCE ANALYSIS}
\date{}

 \author{
  Miriam Louise Carnot\textsuperscript{1}, Eric Peukert\textsuperscript{1}, Bogdan Franczyk\textsuperscript{2}
  }
\address{
	\textsuperscript{1 } ScaDS.AI (University of Leipzig) \\
	\textsuperscript{2 } University of Leipzig and Wroclaw University of Economics\\
	\\
}
\abstract{
In the efforts for safer roads, ensuring adequate vertical clearance above roadways is of great importance. Frequently, trees or other vegetation is growing above the roads, blocking the sight of traffic signs and lights and posing danger to traffic participants.
Accurately estimating this space from simple images proves challenging due to a lack of depth information. This is where LiDAR technology comes into play, a laser scanning sensor that reveals a three-dimensional perspective. 
Thus far, LiDAR point clouds at the street level have mainly been used for applications in the field of autonomous driving. These scans, however, also open up possibilities in urban management.
In this paper, we present a new point cloud algorithm that can automatically detect those parts of the trees that grow over the street and need to be trimmed. 
Our system uses semantic segmentation to filter relevant points and downstream processing steps to create the required volume to be kept clear above the road.
Challenges include obscured stretches of road, the noisy unstructured nature of LiDAR point clouds, and the assessment of the road shape.
The identified points of non-compliant trees can be projected from the point cloud onto images, providing municipalities with a visual aid for dealing with such occurrences. 
By automating this process, municipalities can address potential road space constraints, enhancing safety for all. They may also save valuable time by carrying out the inspections more systematically. Our open-source code gives communities inspiration on how to automate the process themselves.
}
\maketitle
\sloppy

\input{chapters/1_intro}
\input{chapters/2_related_work}
\input{chapters/3_methodology}
\input{chapters/4_results}
\input{chapters/5_discussion}
\input{chapters/6_conclusion}
\input{chapters/acknowledgement}

\bibliography{references}

\end{document}

%% file: chapters/1_intro.tex
\section{Introduction}
\label{sec:intro}

In the domain of urban planning and road infrastructure management, municipalities carry the responsibility of upholding standards for road clearance and maintenance. Besides the immovable elements of urban infrastructure —such as bridges and streetlights— vegetation assumes a particular role. Trees growing over the roadway can not only obscure drivers' view of signs and lanes but also cause damage to vehicles. Falling branches may potentially inflict serious injuries on pedestrians or on bicyclists \cite{way2022}. \\
The vertical clearance above roadways is subject to distinct regulatory frameworks across jurisdictions. Examples include Germany's four-meter standard and the U.S.'s 16-foot specification. In Australia, it varies depending on the type of road. 
Typically, vertical clearance ranges from four to six meters, constituting a crucial factor in ensuring unhindered flow of traffic. \\
Hence, local governments have taken on the task of monitoring the growth of roadside greenery. However, this effort is characterized by its labor-intensive nature necessitating individual assessments for each street. For example, in the municipality of Leipzig, a mid-size German city with around 1,700 kilometers of street, ten workers are employed to regularly inspect the overall road space. An automated approach could not only save resources but also make the roads safer.  \\
Simple cameras are unsuitable for this type of assessment as the resulting images do not contain depth information. Light Detection and Ranging, or LiDAR for short is a sensing technology that measures distances using lasers. With the ability to create a three-dimensional point cloud representation of its environment, LiDAR emerges as the linchpin for this study. \\ 
Working with LiDAR point clouds poses different challenges as also identified in the works of Li et al. \cite{li2020} and Gargoum et al. \cite{gargoum2017}. 
A single point cloud only contains the points that are visible from the location of the sensor, meaning that the LiDAR sensor cannot detect points if there is an object in between.
Also, the range of the sensor is limited. We address this challenge by aggregating multiple point clouds captured at regular intervals, aiming to enhance scene comprehension while avoiding excessively large point clouds that can hinder efficiency.
The efficiency problem persists with the semantic segmentation models which either take a long preprocessing or training time and often require a huge amount of memory. We therefore decided to use RandLANET which reduces computation time by using a random sampling method \cite{hu2020}. \\
To understand where the clearance regulations apply, the outer edges of the road must be found. Finding contours has been dealt with in different point cloud applications even though there is no clear definition of contours in 3D space \cite{xia2020}. Previously proposed methods are based on local features, segments, or gradients. 
We developed a new contour algorithm for 2D planes in 3D space as the other approaches seemed overly complex for our use case.
This algorithm checks the distribution of neighboring points around the point under consideration and then decides whether this point is part of the contour.
Finally, the identified points of the trees that impair the clear height regulations are difficult to locate in the ``real'' world which is why we project those points back to the images.
The contributions of this paper are the following:
\begin{enumerate}
    \item concatenation of multiple point clouds
    \item semantic segmentation for the Pandaset 
    \item a new algorithm for detecting the contours of a plane using nearest neighbors
    \item cropping a sub-point cloud using contours
    \item projecting points from a point cloud onto images
\end{enumerate}
The organization of this study is outlined as follows: Chapter \ref{sec:related_work} surveys existing research in the domains of applications of LiDAR point clouds, and models for semantic segmentation of point clouds. Subsequently, we dive into the chosen dataset and the details of our implementation in Chapter \ref{sec:methodology}. In Chapter \ref{sec:results} we present the results of different experiments we conducted to find the best parameter settings. Chapter \ref{sec:discussion} critically engages with and interprets our approach, describing the practical applications and inherent constraints. Finally, we summarize the key findings and conclusions in Chapter \ref{sec:conclusion}. \\
The code is available at \url{github.com/miriamcarnot/clear_height_pandaset}.

%% file: chapters/2_related_work.tex
\section{Related Work}
\label{sec:related_work}
To develop an approach for the automatic evaluation of the clearance height, we have a look at other applications of LiDAR technology, especially for other clearance problems. We will also briefly discuss semantic segmentation and contour detection in point clouds. 

\subsection{LiDAR applications}
\label{subsec:rw_deapth_measuring}
In the Earth and Geosciences, LiDAR point clouds are mainly taken from the air (remote sensing) to provide an overview of vegetation, its distribution, altitude, etc. \cite{guo2017}. 
From an aerial perspective, it proves challenging to evaluate the clearance above the road which requires data from the ground. \\
LiDAR point clouds captured at street level are primarily recorded for autonomous driving purposes. Autonomous cars should be able to detect objects and estimate distances using LiDAR sensors \cite{li2020}. Beyond driving applications, this data can be used to draw conclusions about roadside vegetation and overall road safety considerations. \\
Yet other studies have also dealt with three-dimensional LiDAR data with regard to traffic and road safety. \\
Gargoum et al. \cite{gargoum2017} summarize applications of LiDAR point clouds in the traffic sector. They report that most of the work in the area to date has been limited to traffic signs and road markings. Yet some studies have already dealt with distance estimates in the context of roads, most of them concerning bridges and tunnels \cite{Watson2012} \cite{puente2016} \cite{liu2012}.
Gouda et al. \cite{gouda2021} developed a novel system for highways that uses raycasting to estimate, among other things, the horizontal roadside clearance, e.g. distances to obstacles or embankments. 
The work of Gargoum et al. focuses on calculating the minimal clearance above highways \cite{Gargoum2018}. They aimed to find out whether over-sized trucks could pass on the selected roads. This is necessary to plan appropriate routes for this kind of transportation. They used LiDAR sensors to detect all objects above three different highway segments. For each trajectory point the algorithm checks if there are any points directly above it. Consecutive points are clustered to determine if the object is a bridge or not. The approach makes sense for the given goal but is not suitable for finding parts of vegetation that need to be trimmed because it only covers the points that are directly above the vehicle.
The work of Chen et al. \cite{chen2019} focuses on the tree inventory along the road. They built a system for identifying individual trees and their specifications using LiDAR sensors. They tested four different initial segmentation approaches including detecting tree clusters but none of them make use of Deep Learning. Most of the Machine Learning models for point clouds were developed in recent years after the publication of this work.

\subsection{Semantic Segmentation of Point Clouds}
\label{subsec:rw_sem_seg}
The semantic segmentation of point clouds using Deep Learning is an active field of research with new models every year. Just like for images where each pixel is assigned a class, for point clouds each point is given a class. In the context of streetscapes, these classes can be pedestrian, car, building, traffic sign, sidewalk, etc.

\subsection{Contour detection in 3D point clouds}
\label{subsec:rw_contours}
In their review, Xia et al. \cite{xia2020} differentiate between three fundamental categories of contour detection algorithms: local-feature-based, segment-based, and gradient-based algorithms. 
The work of Hackel et al. \cite{hackel2016} is based on several different neighborhood features for each point which they use to train a classifier. Neither the model nor the code is available. They labeled contour points to train the network making re-implementation very time-consuming.
Segment-based methods try to separate the point cloud in different areas like planes or triangles and then consider intersection lines as contours. Gradient-based methods usually project the points onto an image or a plane, some of those methods also use voxelization. Xia et al. \cite{xia2017} first detect edges based on geometric centroids, then find candidates by analyzing neighborhood eigenvalues. Finally, they smooth the edges in a linking step with a graph-snapping algorithm. 
Since there is no clear definition for a contour in 3D space, it is difficult to compare different methods quantitatively.

%% file: chapters/3_methodology.tex
\section{Methodology}
\label{sec:methodology}
In this work, we employ point cloud data to identify vegetation encroachment within the vertical clearance above roadways. 
To begin with, we briefly introduce the dataset used.
The first step of the system involves the segmentation of the cloud into semantic classes. Then, we filter the points that belong to the road and approximate the outer road borders by finding the contours of the road plane. The resulting polygon is used for finding vegetation points that lay inside of the vertical clearance height. Finally, the vegetation parts that have grown into the clearance of the road can be projected from the point cloud into the simultaneously acquired images. This not only expedites the process but also helps municipalities to proactively ensure road safety through targeted vegetation management. \\

\subsection{The dataset}
\label{subsec:m_datasets}
Our investigation is conducted utilizing PandaSet\cite{xiao2021}. The dataset originally included 6,080 frames in urban street scenes for semantic segmentation.
The setup includes two LiDAR sensors (360-degree and front-facing) and six cameras taking images simultaneously with the point clouds. \\
We have chosen this dataset due to multiple reasons. Firstly, the developers provide a development kit with tutorials which makes it easy to get started.\footnote{https://github.com/scaleapi/pandaset-devkit} 
Further, the company Hesai providing the dataset also produces the sensors themselves. They are amongst the most affordable on the market and gained popularity in recent years. We are currently planning our own setup with such a sensor and wanted to do a first proof of concept on similar data.
The dataset is currently only available in a shortened form on Kaggle. \footnote{https://www.kaggle.com/datasets/usharengaraju/pandaset-dataset?resource=download}

\subsection{Semantic Segmentation}
\label{subsec:meth_seg}
The point clouds have to be segmented into semantic classes to find which points of the point cloud belong to the vegetation and which points are part of the street. This information is necessary to understand where the height clearance applies at all. As only the labels vegetation and road are relevant for this work, we trained the RandLANet model only on three classes: the two above-mentioned and other objects.
The shortened dataset includes semantic labels for 30 sequences of point clouds, each depicting a different scene. 
We decided to train a RandLANet model as it is often referred to as a baseline and keeps the initial preprocessing at a minimum by using random down sampling. The authors have shown in their paper that more sophisticated sampling methods do not yield big advantages and are very time-consuming compared to the random approach.  \\
We trained the model on 24 sequences for 75 epochs, four were left for testing and two for validation.
For training and visualization, we use Open3D-ML, a machine learning library built on top of the 3D data processing library Open3D \cite{Open3D-ML}\cite{zhou2018}. It offers ready pipelines for different tasks such as semantic segmentation and unifies the definition of datasets, configurations, and models.  

\subsection{Concatenation of Point Clouds}
\label{subsec:concatenation}
\begin{figure}[H]
    \centering
    \subfloat[]{\includegraphics[width=0.40\textwidth]{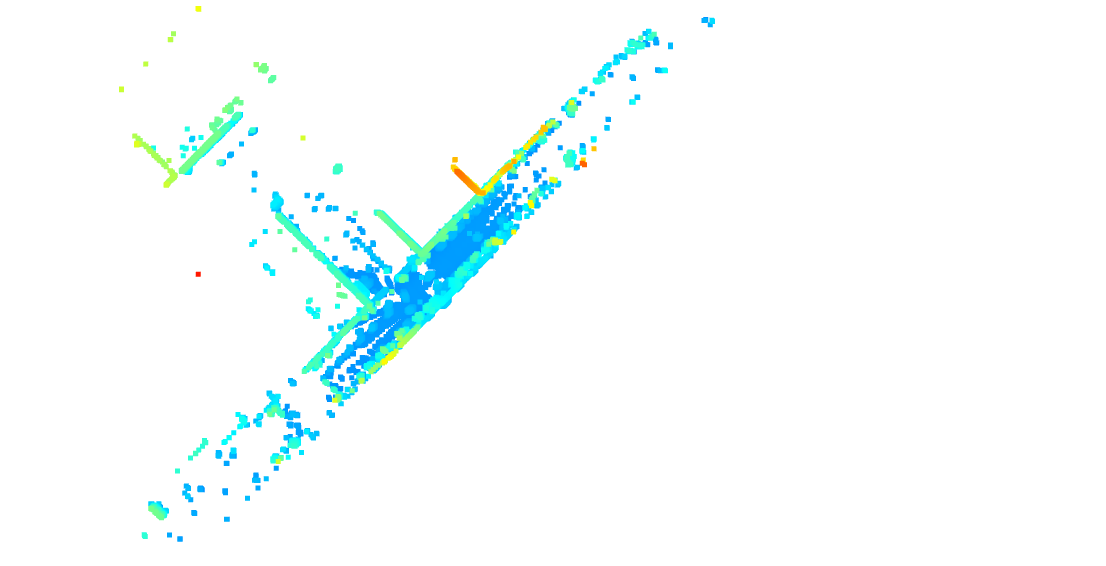}}\hfill
    \subfloat[]{\includegraphics[width=0.40\textwidth]{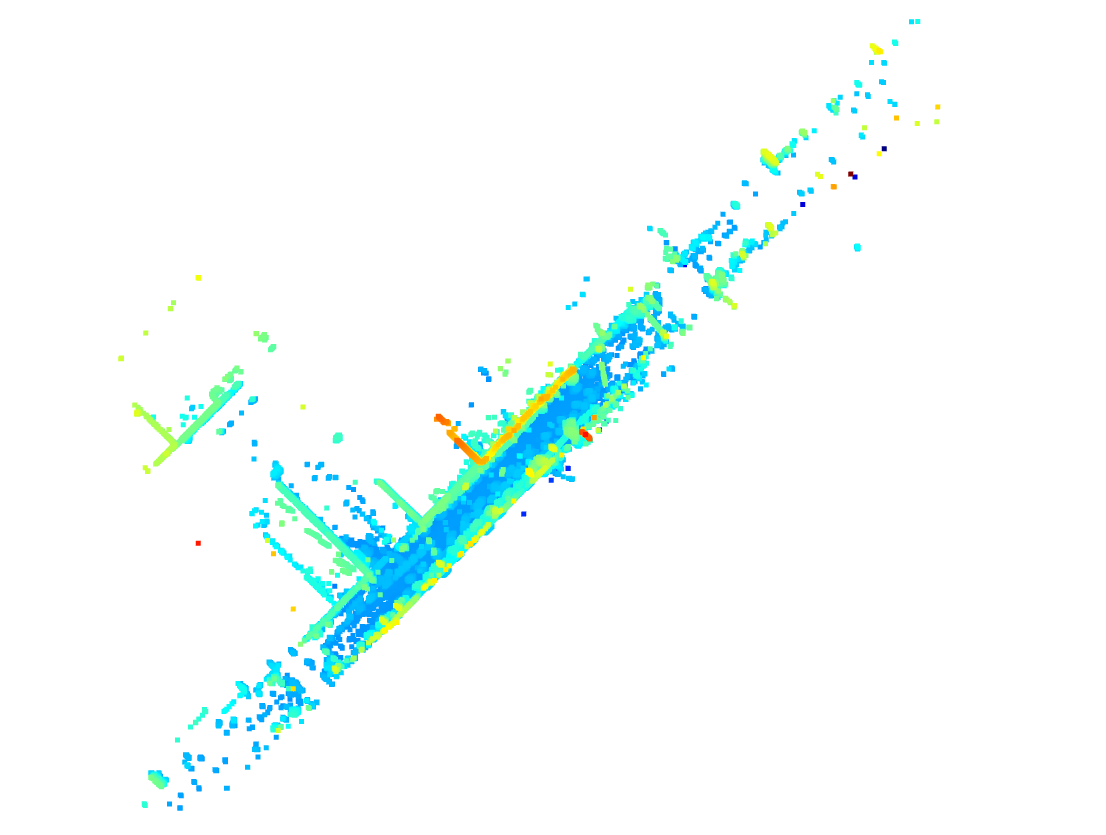}}
    \caption{Single (a) and concatenated (b) point clouds of sequence 003.}
    \label{fig:concatenation_of_pcds}
\end{figure}

One point cloud is only showing one recording moment where many objects can be blocked by other objects. Therefore, we concatenate multiple point clouds of one sequence to have a better overview of the entire street scene. To not work with too many points which enlarges the execution time we take every i-th point cloud of a sequence. The decision of the concatenation step i will be discussed in Section \ref{subsec:res_concat_step}. As is shown in figure \ref{fig:concatenation_of_pcds} more detail of the street scene can be observed when concatenating several recordings, here with a step of 10.

\subsection{Edge Detection of the Road Plane}
\label{subsec:edge_detection}
\begin{figure}[H]
\centering
\includegraphics[width=0.9\linewidth]{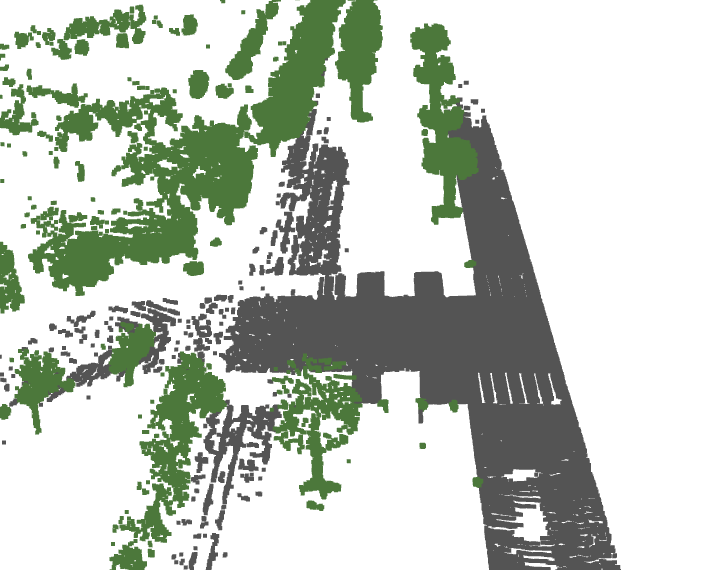}
\caption{Point cloud containing only points belonging to the classes road (grey) and vegetation (green) in sequence 011.}
\label{fig:filtered_pcd}
\end{figure}

\begin{figure}[H]
    \centering
    \subfloat[]{\includegraphics[width=0.45\linewidth]{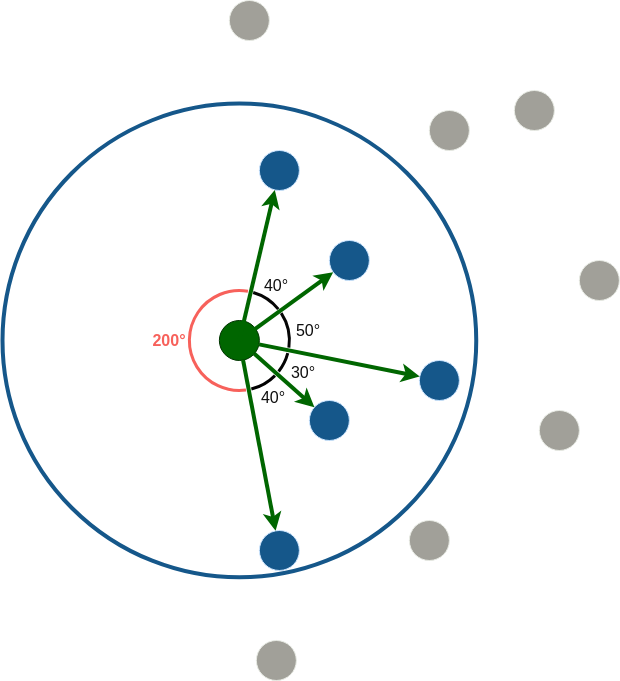}} \hfill
    \subfloat[]{\includegraphics[width=0.5\linewidth]{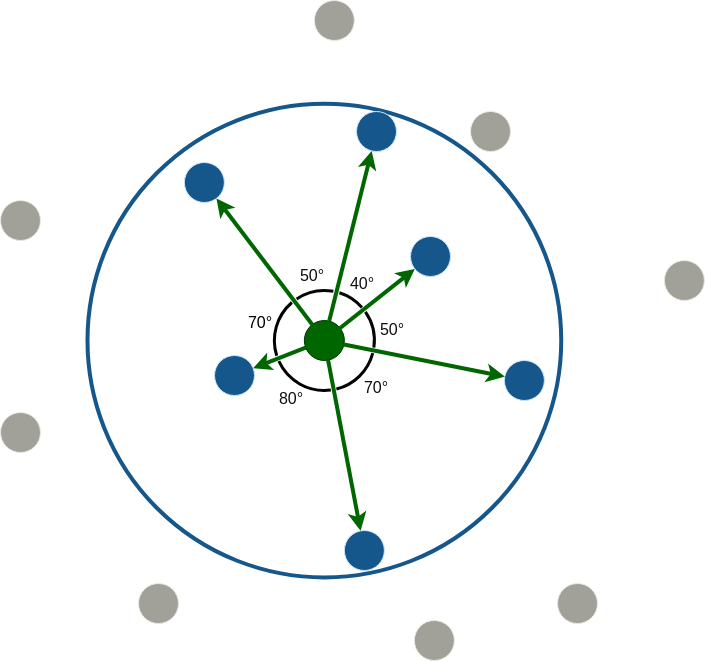}}
    \caption{The green point in a) is considered a contour point because one angle exceeds 135° (shown in red). The green point in b) is not a contour point.}
    \label{fig:contour_algorithm}
\end{figure}

Once the point cloud is segmented, we can continue working only on the classes that are interesting for our use case which are vegetation and road. Figure \ref{fig:filtered_pcd} shows an example of a point cloud that has been filtered for those two classes. \\
The next step is to find out where exactly the road ends, therefore, we have to find the contour of the road. The road points roughly form a plane thus making the task less complex as the third dimension can be left out.
Other reviewed research projects try to detect edges and contours in true three-dimensional objects such as buildings. These methods seem to be overly sophisticated in our case often requiring extensive calculations or the training of models. We thus developed a new basic approach to check if a point belongs to the contour. \\
The execution of this algorithm for each road point requires a considerable amount of computing power. In a preprocessing step, we therefore use Poisson sampling to reduce the road points to a smaller number of sampling points as representation. For a given point cloud, points are gradually eliminated to ensure that the distances between the points are as uniform as possible.
We execute the later program for each point of the down-sampled cloud in parallel to speed up the process. \\
Figure \ref{fig:contour_algorithm} shows the main idea of our developed algorithm to find out whether a point belongs to the contour. Our algorithm checks every sampling point of the road.
For each sampling point \(P\) we calculate all the neighboring street points within a radius \(r\). Let \( N \) be the set of all neighboring street points, defined as: 
\begin{equation}
N = \{ N_i \mid \text{distance}(P, N_i) \leq r \}
\end{equation}
If there are no other points close \(P\) is not considered a contour point. If there is only one neighbor \(P\) is automatically accepted. \\
For each neighboring point \( N_i \), we calculate a vector \(V_i\) between \(P\) and \(N_i\) ignoring the z-coordinate as the street is considered a plane. x and y are the remaining two coordinates of the points. 
\begin{equation}
V_i = \begin{bmatrix} N_{i,x} - P_x \\ N_{i,y} - P_y \end{bmatrix}
\end{equation}
Then, we start at any vector \(V_i\) and calculate the clock-wise angle to every other vector \(V_j\) and choose the smallest one:
\begin{equation}
\theta_{\text{min}} = \min_{i,j \in N}\left(\text{atan2}\left(\begin{bmatrix} V_{i,x} \\ V_{i,y} \end{bmatrix}, \begin{bmatrix} V_{j,x} \\ V_{j,y} \end{bmatrix}\right)\right)
\end{equation}
We continue from the vector \(V_j\) that had the smallest angle with \(V_i\) and continue this procedure until all of the vectors have been checked and the sum of the smallest angles adds up to 360°. \\
In the case that the neighboring points \(N\) are well distributed around \(P\), none of those angles will be of significant size and exceed the threshold as can be seen in figure \ref{fig:contour_algorithm} b). The point \(P\) is not considered a contour point in this case. If one of the angles is very large, then \(P\) is added to the list of contour points. A more detailed analysis of the determination of the angle threshold value follows in section \ref{subsec:res_angle_thresh}.  
\[
\text{Contour Point} = \begin{cases} 
1 & \text{if } \text{min\_angle} > threshold \\
0 & \text{otherwise}
\end{cases}
\]\\
\begin{figure}
    \centering
    \subfloat[]{\includegraphics[width=0.3\linewidth]{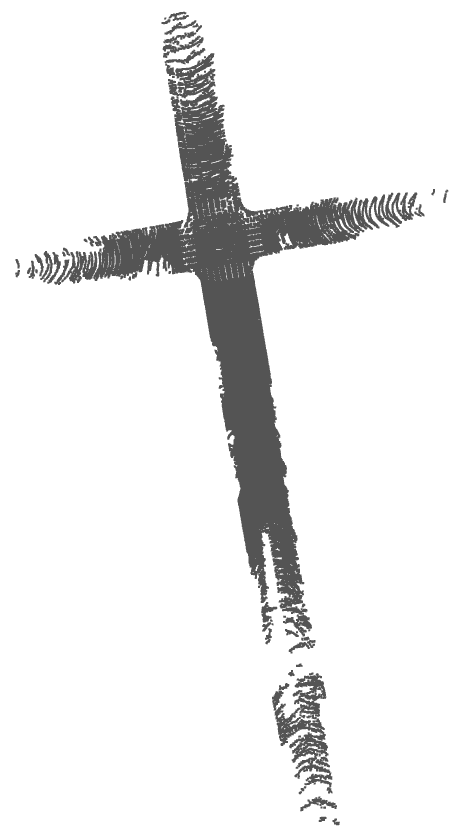}}
    \subfloat[]{\includegraphics[width=0.3\linewidth]{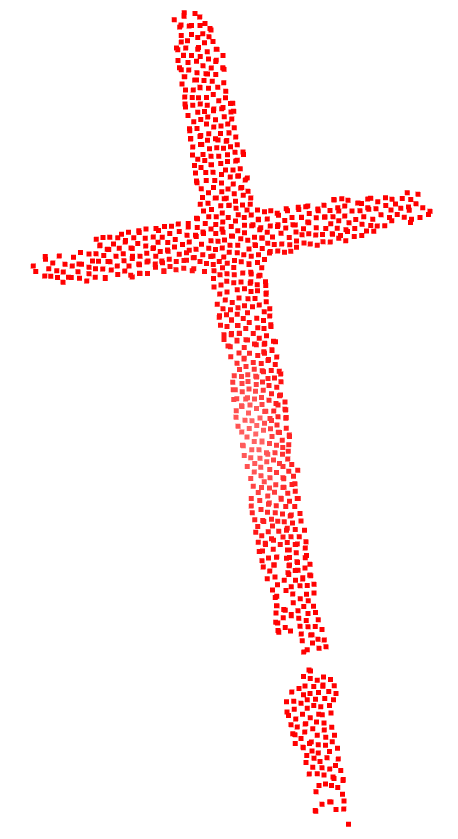}}
    \subfloat[]{\includegraphics[width=0.3\linewidth]{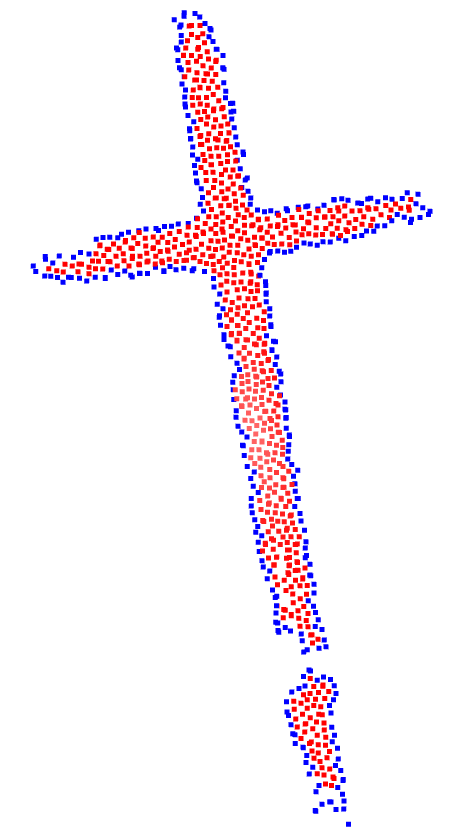}}
    \caption{All the road points (a), the sampled road points (b), and the detected contour points in blue.}
    \label{fig:contour_example}
\end{figure}
Figure \ref{fig:contour_example} shows an example of the contour-finding process using sequence 027 of the dataset. Figure \ref{fig:contour_example} a) includes all points that belong to the road class after removing the statistical outliers, in total 299,599. The following step samples the points to 1,000 representing points, shown in b). Finally, we check for each point if it is part of the contour. Contour points are shown in blue, others remain red as can be seen in c).

\subsection{Spanning the Clearance Gauge}
\label{subsec:spanning_cl_gauge}

\begin{figure}[hbt!]
\centering
\includegraphics[width=\linewidth]{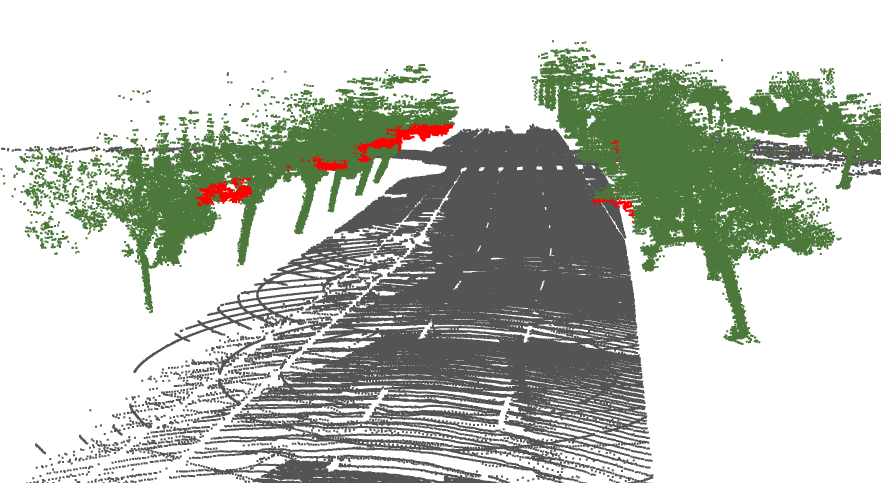}
\caption{The detected points are shown in red (sequence 019).}
\label{fig:result}
\end{figure}

Having the contour points, we can connect all of them to a polygon. For this purpose, we sort the contour points according to their distance from each other. We start at any random contour point and look for the closest point. We continue this procedure until we have sorted all the contour points. We then form a line set that includes all the lines connecting neighboring contour points representing the polygon. \\ 
Using this line set, we can create a volume that ranges from the street level up to four meters which is the height clearance regulation in Germany. The created volume is used to filter out points from the vegetation class that lay inside this volume. These are the points of the trees that need to be trimmed. In Figure \ref{fig:result} these points are marked in red. \\
For working with and visualizing the point clouds, we use the library Open3D \cite{zhou2018}, which makes working with 3D data an interactive experience.

\subsection{Projection onto Images}
\label{subsec:projection}

\begin{figure*}[hbt!]
  \centering
  \includegraphics[width=\textwidth]{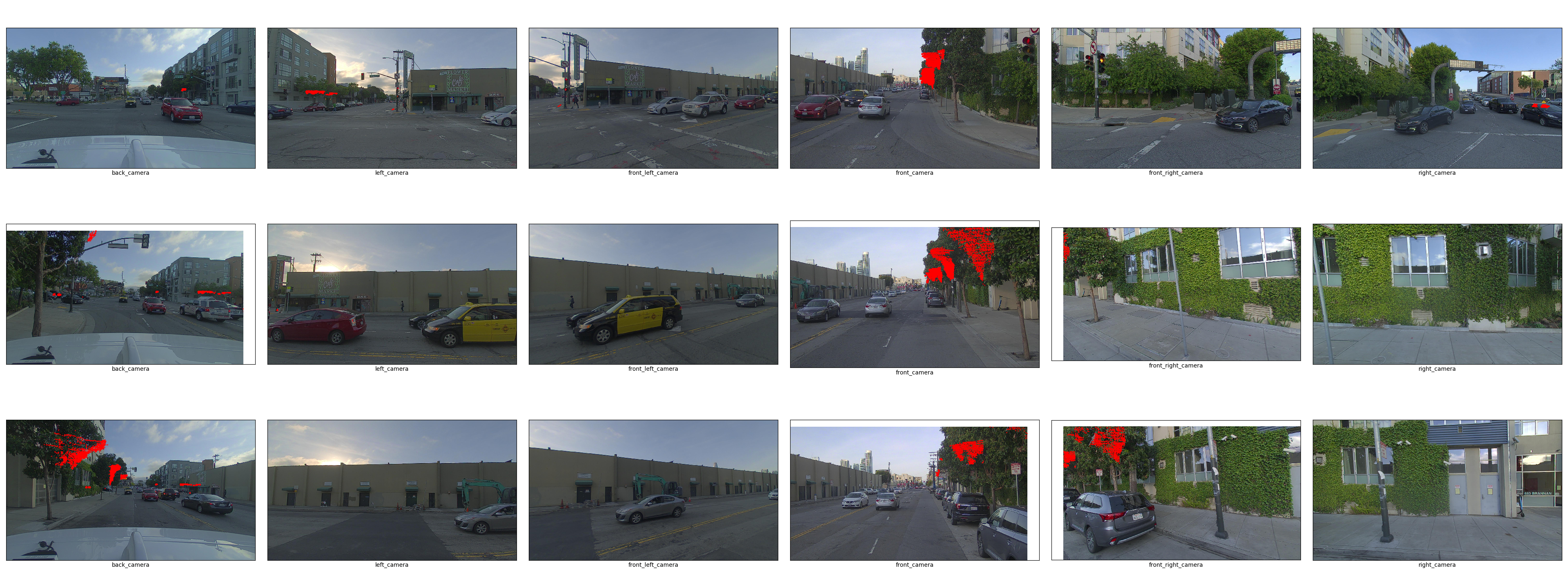}
  \caption{Images from three different time steps including the projected points (sequence 001).}
  \label{fig:projection}
\end{figure*}

Besides the LiDAR sensor, the car recording the Pandaset was equipped with six cameras in different directions. The points that compromise the clear height above the road can be projected onto those images using the information about the position of the car when the images were taken. The result of this projection for three different steps (rows) and the six cameras (columns) are presented in Figure \ref{fig:projection}.

%% file: chapters/4_results.tex
\section{Results}
\label{sec:results}

The following section describes the conduct and results of several experiments. The aim is to identify good default values for the parameters taking into account the execution time, the quality of the contours, and the number of points on trees to be trimmed.
As already mentioned in Section \ref{sec:related_work}, there are to our best knowledge no other algorithms with the same objective. This is why we cannot directly compare the implemented system to others. Instead, we decided to test the setting of the following parameters:
\begin{itemize}
    \item point cloud concatenation step 
    \item number of sampling points 
    \item neighborhood radius for the contour algorithm 
    \item angle threshold for the contour algorithm 
\end{itemize}
We compare the qualitative results in the form of images and the computing time required, when appropriate. \\
Finally, we will compare the annotated labels with the predictions from the semantic segmentation model. 

\subsection{Point Cloud Concatenation Step}
\label{subsec:res_concat_step}
To ensure a comprehensive depiction of the scene, we select every \(i\)-th point cloud of a sequence (which consists of 80 point clouds in total) and concatenate them. The objective of the experiment is to determine the optimal parameter for \(i\). Concatenation steps 1, 5, 10, 20, and 40 are employed for this purpose. The experiment is repeated for each sequence, and the averages of the obtained values are utilized for fair comparison. \\
Table \ref{tab:concat_step} presents the outcomes of the experiment. Naturally, a reduced concatenation step leads to a greater number of point clouds and a proportionally higher number of points. Additionally, the concatenation process time exhibits no significant escalation with smaller concatenation steps. However, the overall program execution time markedly increases when employing a smaller concatenation step due to the inclusion of more frames. With a higher total number of points, the number of vegetation points within the clear height also increases, thereby offering enhanced detail in areas requiring tree pruning.
Upon careful consideration, we have adopted the use of every 10th frame as the default, as it creates a balanced relationship between the execution time and the level of detail achieved.
\begin{table*}
\caption{The table shows the tested values for the concatenation steps, the resulting number of point clouds that are being concatenated, and the total number of points. Further, we calculate the average number of vegetation points that lay inside the clear height, and the average time it takes to concatenate the frames and execute the entire program in seconds.}
\centering\small
\begin{tabular}{c c c c c c}
\toprule
Concatenation Step & \# Point Clouds & \# Points & Vegetation Inliers & Concatenation Time in s & Execution Time in s \\
\hline
1   &   80  &   13,823,260  &   129,281   &   0.572   &   100.911 \\
5   &   16  &   2,764,037   &   40,266    &   0.242   &   19.700  \\
10  &   8   &   1,383,692   &   12,421    &   0.205   &   10.636   \\  
20  &   4   &   693,050     &   9,238     &   0.186   &   6.649   \\
40  &   2   &   346,436     &   9,184     &   0.187   &   5.703   \\
\bottomrule
\end{tabular}
\label{tab:concat_step}
\end{table*}

\subsection{Number of Sampling Points}
\label{subsec:res_sampling_points}
Similar to the previous parameter, setting of the number of sampling points involves a balance between precision and runtime considerations. We systematically examined the values of 100, 250, 500, 1000, and 2000, conducting each experiment for every sequence and subsequently computing the average. \\
Table \ref{tab:sampling_number} displays the outcomes, illustrating that the sampling process significantly contributes to the overall execution time. This is especially the case for low numbers of sampling points like 100 or 250, where the sampling process takes up about 90\% of the execution time of the entire program. With increasing numbers of sampling points, most time is needed for the edge detection of the road. \\ 
As anticipated, the quantity of vegetation points within the clear height exhibits an upward trend with an increasing number of sampling points, attributed to the heightened detail captured in the contours. The high number of sampling points facilitates a more accurate determination of the road's shape, ensuring comprehensive coverage without omission of any sections. 
\begin{table}[H]
\caption{The table shows for different numbers of sampling points how many vegetation points lay inside the clear height and how much time in seconds it takes to sample the road points and execute the entire program.}
\centering\small
\begin{tabular}{c c c c}
\toprule
\# Sampling     & Avg Vegetation    & Sampling  & Execution \\
Points          & Inliers           & Time in s & Time in s\\
\hline
100   &   27,903    &   7.298   &   8.041 \\
250   &   25,459    &   7.308   &   8.042  \\
500   &   21,307    &   7.338   &   8.403   \\  
1000  &   17,359    &   7.498   &   10.864   \\
2000  &   24,642    &   7.828   &   32.503   \\
\bottomrule
\end{tabular}
\label{tab:sampling_number}
\end{table}
\begin{figure}
\centering
\includegraphics[width=0.7\linewidth]{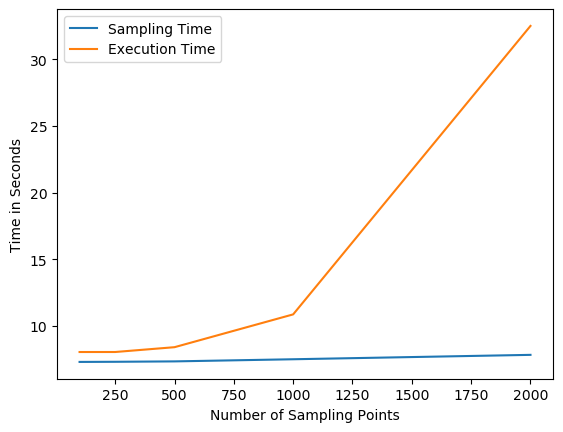}
\caption{Time in seconds for sampling the road points (blue) and executing the entire program (orange).}
\label{fig:sampling_time}
\end{figure}
The corresponding plot in Figure \ref{fig:sampling_time} depicts the execution times, revealing that the time for sampling points exhibits minimal variation, while the overall program execution time experiences a pronounced surge with a greater count of sampling points. This is a consequence of checking individually for each sample point whether it is part of the contour. \\
Figure \ref{fig:sampling_example} graphically depicts the implications of varying the number of sampling points. It illustrates that an increased number of points used as samples enhances the fidelity of the representation of the original point cloud. In this example, the road point cloud comprises 316,574 points after the removal of statistical outliers (depicted in Figure 8a). As a default, we decided on 1000 sampling points. However, the complexity of the street scene should be considered when setting the parameter. 
\begin{figure}[H]
    \centering
    \subfloat[]{\includegraphics[width=0.15\textwidth]{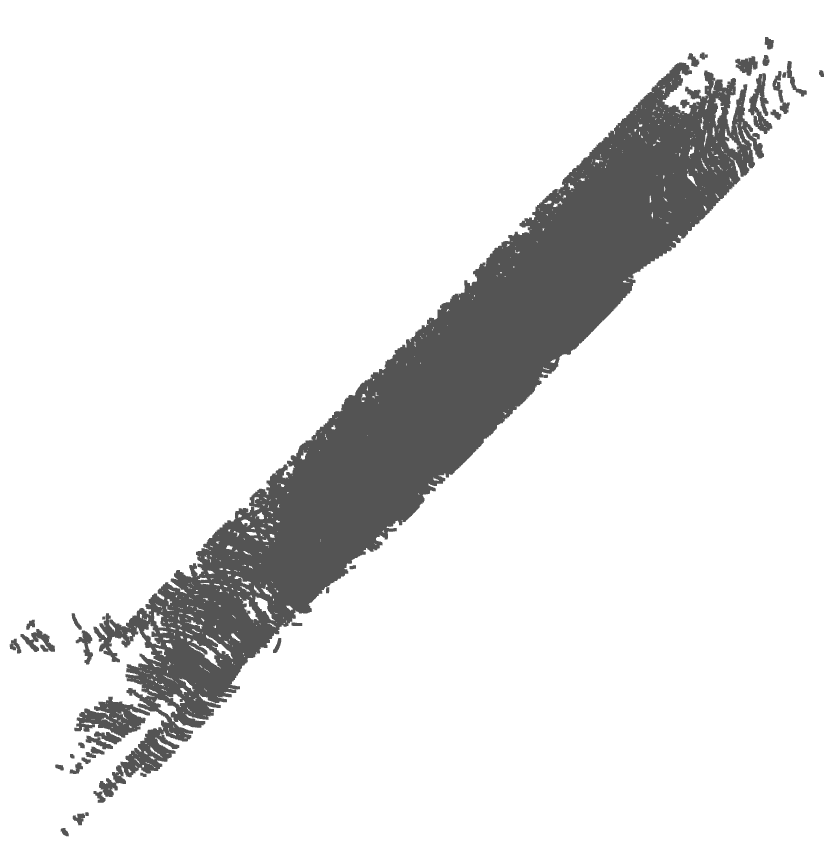}}
    \subfloat[]{\includegraphics[width=0.15\textwidth]{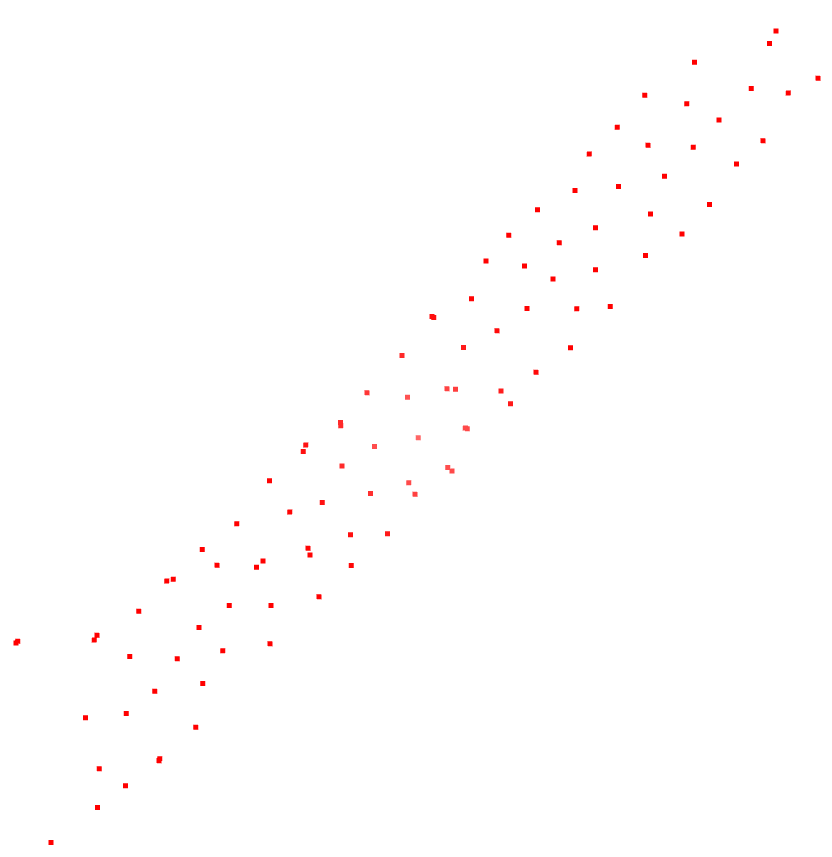}}
    \subfloat[]{\includegraphics[width=0.15\textwidth]{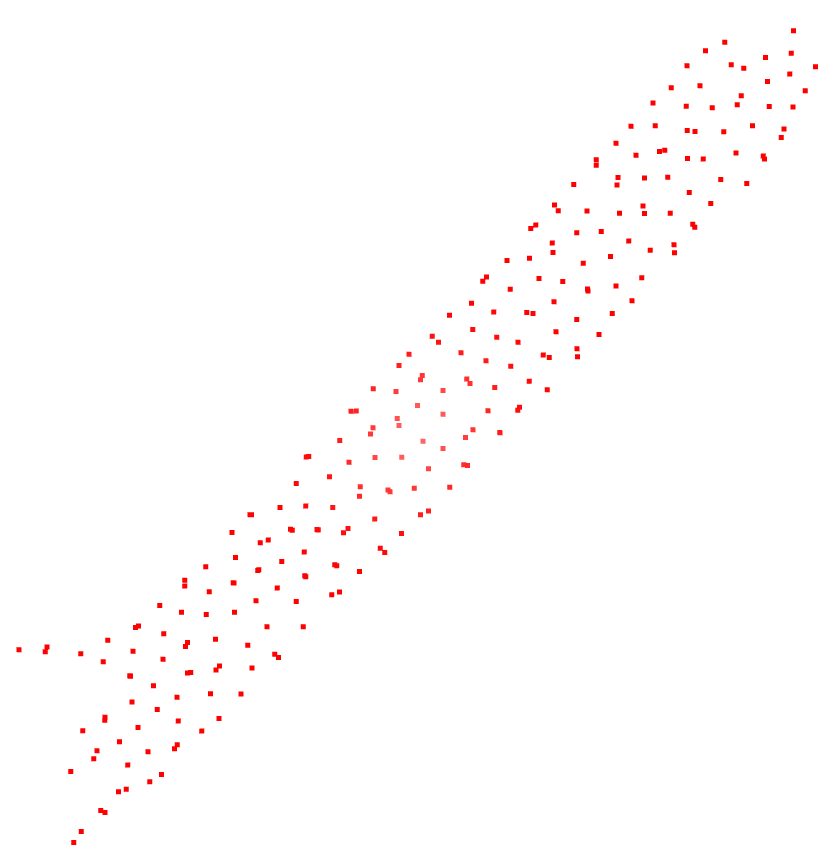}}\hfill
    \subfloat[]{\includegraphics[width=0.15\textwidth]{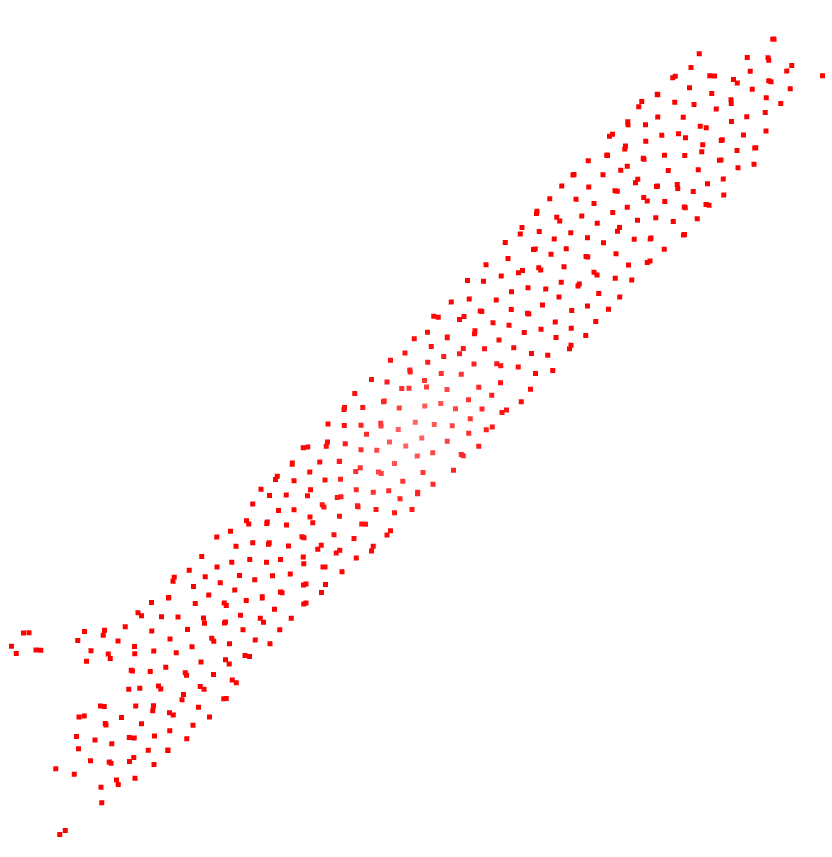}}
    \subfloat[]{\includegraphics[width=0.15\textwidth]{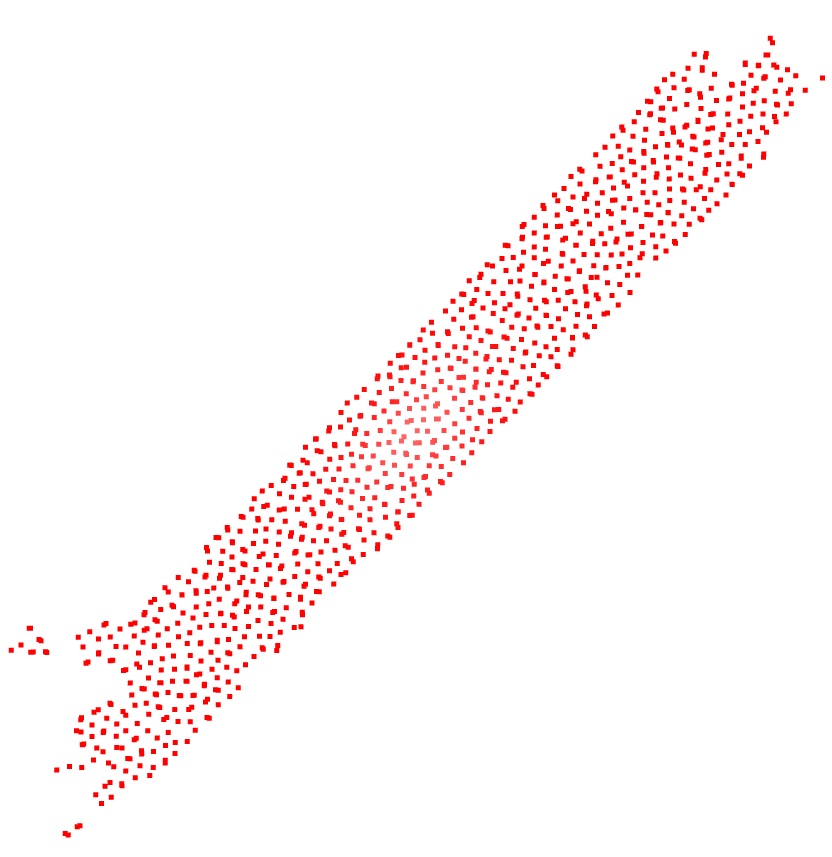}}
    \subfloat[]{\includegraphics[width=0.15\textwidth]{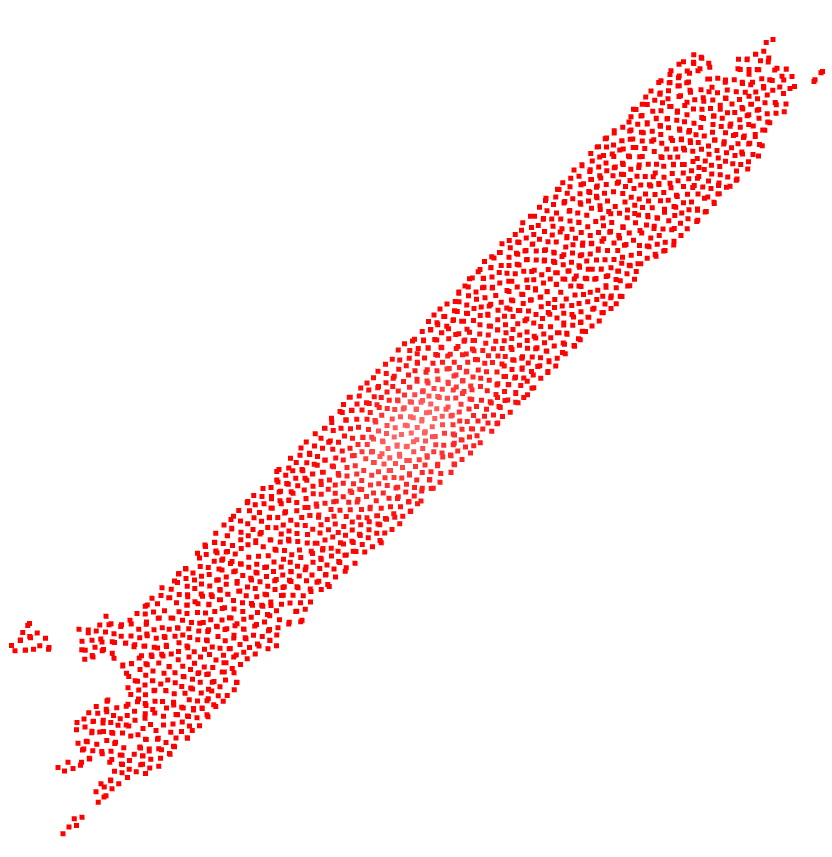}}
    \caption{Road point cloud after removing outliers (a) and sampled road point cloud with 100 (b), 250 (c), 500 (d), 1000 (e), and 2000 (f) points, using sequence 003.}
    \label{fig:sampling_example}
\end{figure}
\begin{table}[H]
\caption{The execution time of the contour algorithm and the entire program in seconds for different sizes of the neighborhood radius.}
\centering\small
\begin{tabular}{c c c}
\toprule
Radius     & Time to Find        & Execution \\
           & Contours in s      & Time in s  \\
\hline
2               &     0.284         &   8.800   \\
4               &     0.593         &   8.625   \\
6               &     2.054         &   10.130   \\  
8               &     6.208         &   14.304  \\
10              &     12.723        &   20.837  \\
\bottomrule
\end{tabular}
\label{tab:radius}
\end{table}

\subsection{Neighborhood Radius}
\label{subsec:res_neighborhood_radius}
\begin{figure}[H]
    \centering
    \subfloat[]{\includegraphics[width=0.15\textwidth]{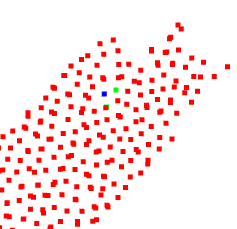}}
    \subfloat[]{\includegraphics[width=0.15\textwidth]{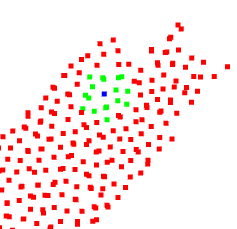}}
    \subfloat[]{\includegraphics[width=0.15\textwidth]{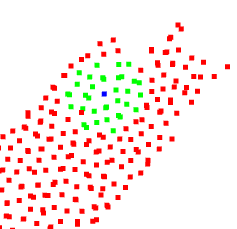}}\hfill
    \subfloat[]{\includegraphics[width=0.15\textwidth]{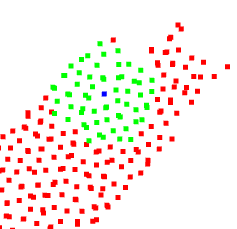}}
    \subfloat[]{\includegraphics[width=0.15\textwidth]{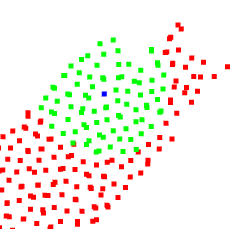}}
    \caption{Neighborhood points (green) of the query point blue with radius sizes 2 (a), 4 (b), 6 (c), 8 (d), 10 (e).}
    \label{fig:radius_example}
\end{figure}

To determine whether a point qualifies as a contour point, an examination of its neighborhood is conducted, defined by a designated neighborhood radius. A smaller radius accelerates execution by considering fewer neighbors; however, there is a risk of omitting crucial neighbor points, leading to potential misclassification. The experiment assesses radius settings of 2, 4, 6, 8, and 10, demonstrating in Table \ref{tab:radius} the corresponding execution times. For the contour algorithm as well as for the full program the execution times rise with increasing radii. The impact of varying radius sizes on the resulting neighborhood is exemplified in Figure \ref{fig:radius_example}. The point in question is marked blue and its neighbors in green.
In this particular example, a radius of 2 yields 2 neighboring points, whereas a radius of 4 encompasses 18, and a radius of 6 comprises 36 neighbors. Larger radii, such as 8 and 10, consider 63 and 88 neighbors, respectively. Despite the rise in execution time with increased neighbors, there is no corresponding enhancement in the informative value of the neighborhood concerning the contour. Only the closest neighbors are truly relevant for the decision of whether the point is a contour. A too-big radius might even have negative implications in complex street environments with multiple sections. 
Therefore, a radius of six was selected as the default size, offering a balance between computational efficiency and the informative relevance of the neighborhood in relation to the contour.

\subsection{Angle Threshold}
\label{subsec:res_angle_thresh}
The angle threshold is another crucial parameter in discerning whether a point is part of the contour. If it is set too small, too many points will be considered as a contour point. If it is too big, we do not find them all. 
We systematically examine angle thresholds of 60°, 90°, 120°, and 150°, as demonstrated in Figure \ref{fig:angle_example}, where the identified contour points are depicted in blue and the remaining sampled road points in red.
For example, setting the angle threshold to 60° (Figure 10a) leads to the classification of numerous points belonging to the ``second'' row as contour points, an undesired outcome. Conversely, a high angle threshold, such as 150° (Figure 10d), fails to identify some genuine contour points. Striking a balance, we propose 90°, as illustrated in sub-figure b, as an optimal value and have adopted it as the default setting. \\

\subsection{Semantic Segmentation on the Pandaset}
\label{subsec:res_sem_seg}

\begin{figure}[H]
    \centering
    \subfloat[Annotation]{\includegraphics[width=0.4\textwidth]{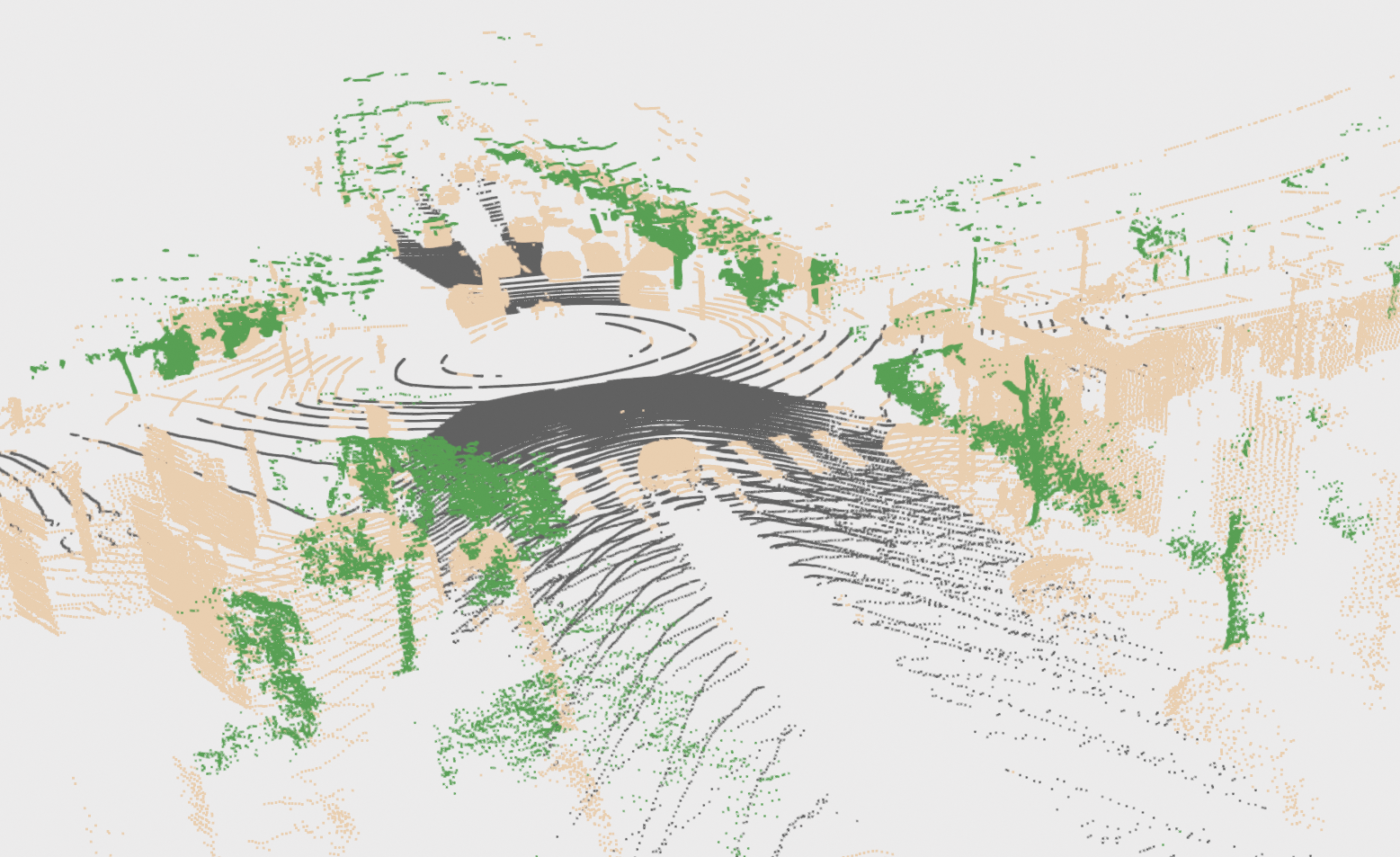}} \hfill
    \subfloat[Prediction]{\includegraphics[width=0.4\textwidth]{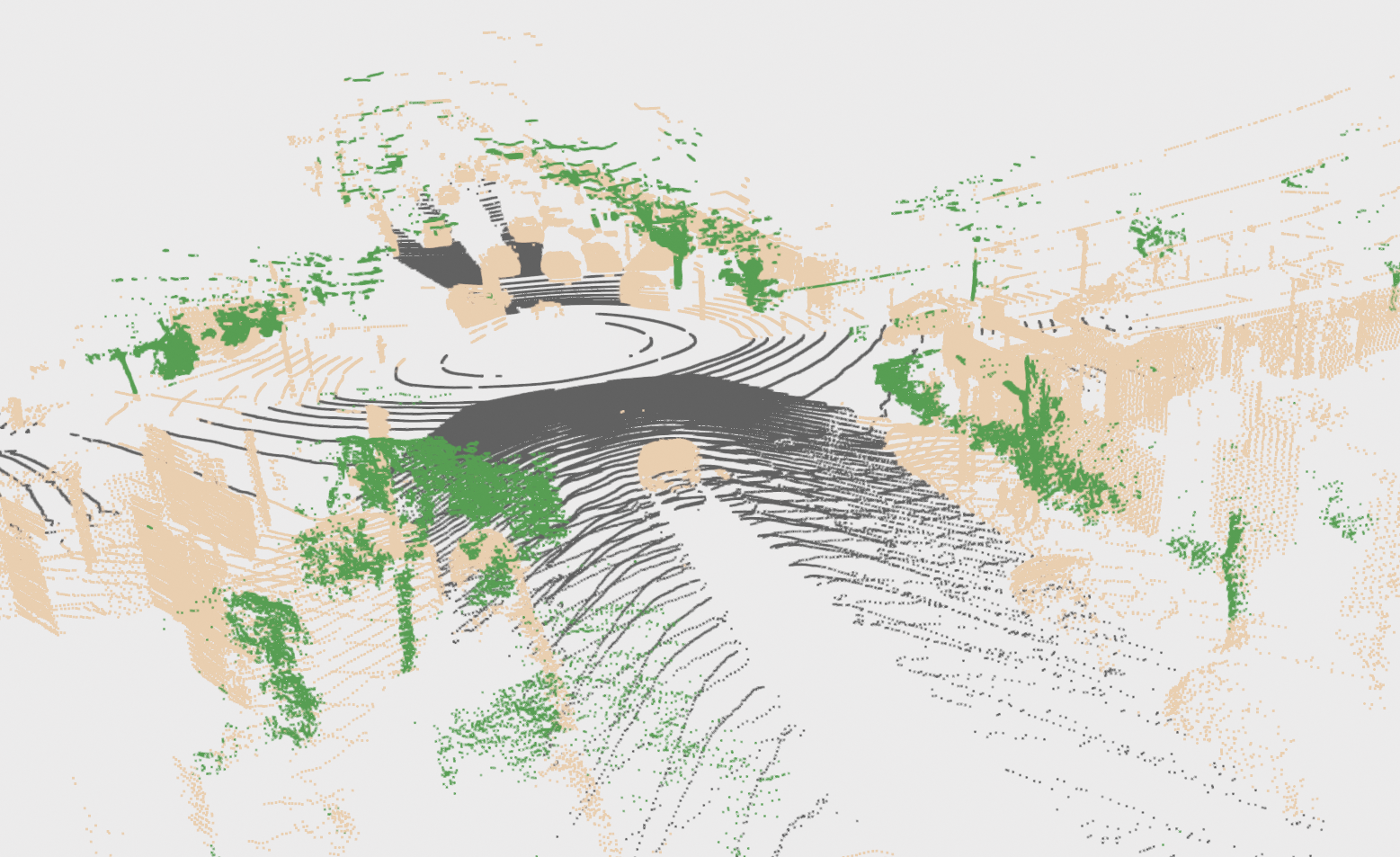}}
    \caption{Annotation and model prediction for a test point cloud. Vegetation is shown in green, roads in grey, and other classes in beige.}
    \label{fig:comp_labels}
\end{figure}

\begin{figure*}
    \centering
    \subfloat[341 contour points]{\includegraphics[width=0.25\textwidth]{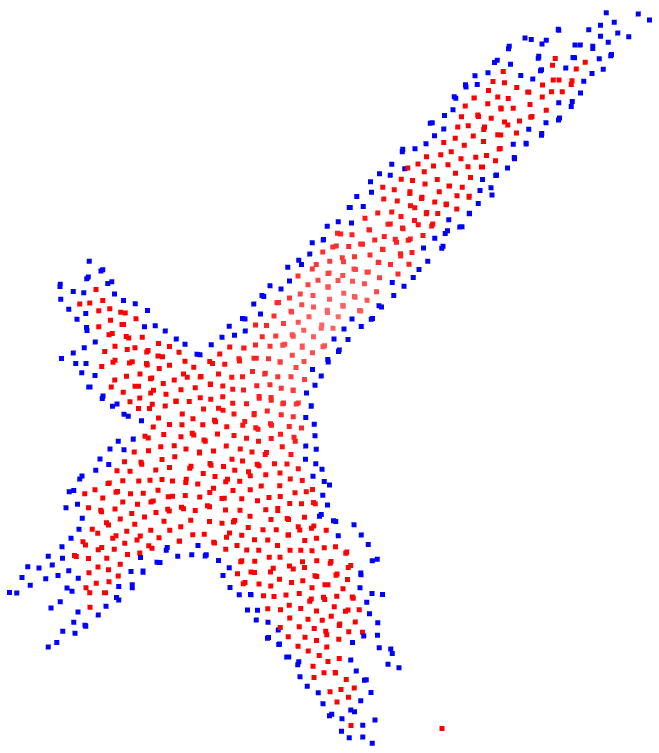}}
    \subfloat[246 contour points]{\includegraphics[width=0.25\textwidth]{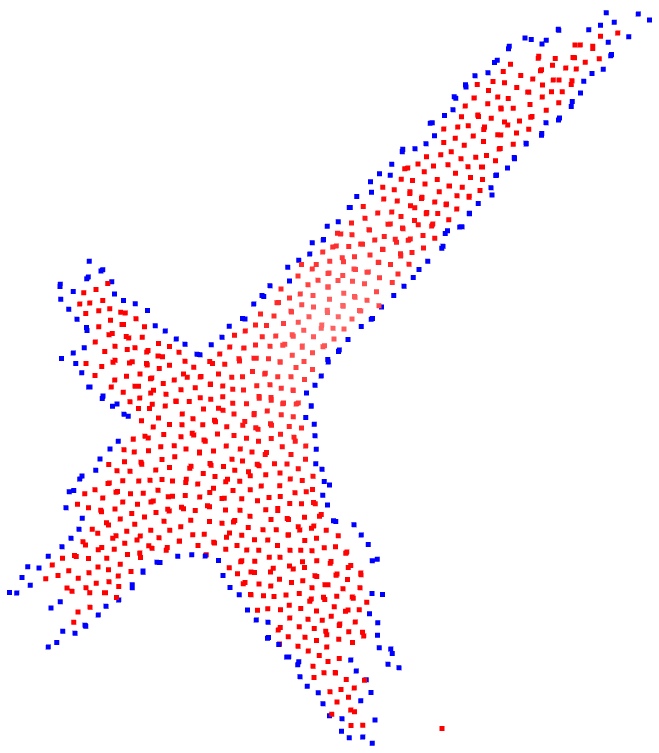}}
    \subfloat[198 contour points]{\includegraphics[width=0.25\textwidth]{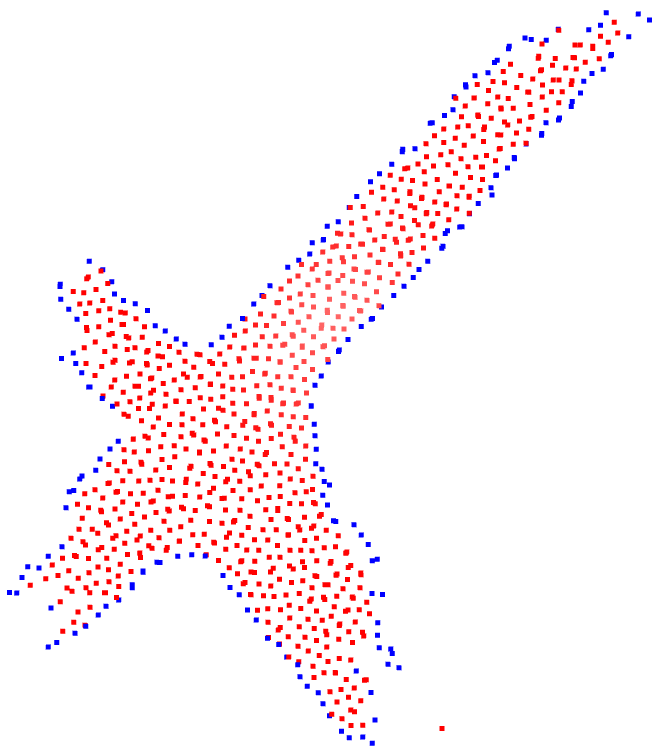}}
    \subfloat[142 contour points]{\includegraphics[width=0.25\textwidth]{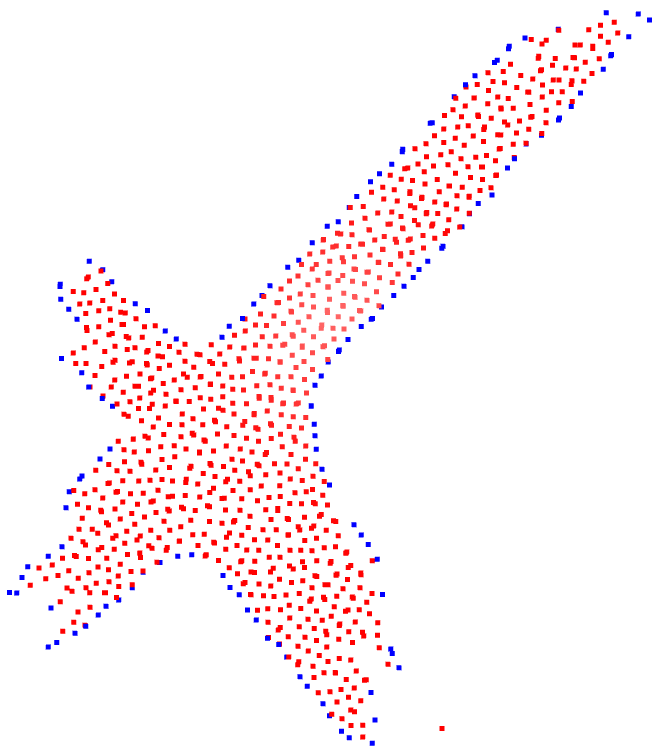}}
    \caption{Detected contours (blue) of the sampled road points (red) with angle thresholds 60° (a), 90° (b), 120° (c), 150° (d).}
    \label{fig:angle_example}
\end{figure*}

We trained the model according to the description in Section \ref{subsec:meth_seg} reaching an average accuracy of 93.86\%. In total 320 point clouds (from four sequences) were being tested. The single accuracies range from 85\% to 97\%. As can be seen in Figure \ref{fig:comp_labels}, the predictions are very close to the annotated labels. In our opinion, a significant part of the accuracy loss is due to the road markings not being considered as part of the road in the dataset. Our model classifies these road marking points as road points, which lowers the accuracy but is positive for our case. This behavior is also visible in the Figure. \\
Table \ref{tab:comp_annos_pred} shows how long it takes to execute the entire program when simply using the annotations compared to using the predictions from the trained model. It is apparent that running the inference for predicting the labels, reduces the speed drastically. For every of the four test sequences, the program takes more than double the time to execute.

\begin{table}[H]
\caption{The table shows the execution time in seconds for the four test sequences when using the given annotations compared to using the model predictions.}
\centering\small
\begin{tabular}{c c c}
\toprule
      & Execution Time in s & Execution Time in s \\
seq   & With Annotations    & With Model Predictions \\
\hline
013   &   10.387  &   32.270       \\
027   &   15.102  &   37.368       \\
029   &   19.974  &   43.429       \\  
046   &   15.670  &   36.856       \\
\bottomrule
\end{tabular}
\label{tab:comp_annos_pred}
\end{table}

%% file: chapters/5_discussion.tex
\section{Discussion}
\label{sec:discussion}

LiDAR point clouds offer substantial advantages for urban information retrieval, yet current research and processing capabilities remain in the preliminary stages. Notably, the variability in LiDAR point clouds, arising from differences in sensor characteristics (e.g., channel count, resolution) and post-processing methods, poses a challenge to the applicability of segmentation models across diverse point cloud types. 
Moreover, the existing labeling tools are numerous but often exhibit only limited functionality. Specifically, for addressing the clear height issue, the labeling tool should be able to measure distances or at least bounding box dimensions and have the ability to visualize original labels for creating subgroups of the semantic classes, in our case vegetation in need of trimming. None of the open-source tools tested thus far provided us with both features. \\
With several separate road sections, for example, double lanes, it can happen that the contour points cannot be connected correctly. The algorithm should be extended so that separate sections are automatically recognized and considered separately.
Although this was not the case in the dataset, it is quite conceivable that the algorithm could run into problems if the gradient of the road varies greatly in a road segment, as the road itself is approximated as a flat surface.
We are actively addressing these challenges in our ongoing work for refinement. \\
To scale the algorithm to point clouds of an entire city, the overall point cloud would have to be divided into segments, which would then need to be analyzed separately. We can use an example to approximate the computing time. Leipzig is a medium-sized German city with just under 600,000 inhabitants and around 1,700 kilometers of road. According to our calculations using the coordinates of the recording points, the Pandaset sequences have a total length of approximately 2.192 meters and can be evaluated within 10.5 minutes with the default settings.
If the entire city of Leipzig were to be recorded with the setup used to create the panda set, the evaluation of the clearance height would take about 136 hours with standard parameter settings. In this respect, further efforts may be undertaken to reduce the processing time.

%% file: chapters/6_conclusion.tex
\section{Conclusion}
\label{sec:conclusion}

Trees growing along streets can embellish the cityscape but frequently need to be taken care of. In this work, we have shown the possibility of automating the process of detecting trees growing over the street and therefore penetrating clear height regulations. We successfully show that parts of trees in need of trimming can be identified within 3D point clouds. \\
One drawback of LiDAR point clouds is their size, recordings can easily become very large. We still believe that LiDAR has a lot of potential in this context, especially with the rise of autonomous driving using LiDAR sensors. Applications for a smart city could be linked to such developments, for example by detecting problems in the road space automatically in the background and sending information to the officials. \\
The overall goal of our project is to equip garbage trucks that regularly cover the entire road network in a city with LiDAR sensors. Collected data can be used for automatic identification of restrictions on clearance height. \\
We hope that our work can give inspiration to cities and communities to save workforce and digitize our cities to pave the way for the future.

%% file: chapters/acknowledgement.tex
\section*{Acknowledgement}

The authors acknowledge the financial support by the Federal Ministry of Education and Research of Germany and by Sächsische Staatsministerium für Wissenschaft, Kultur und Tourismus in the programme Center of Excellence for AI-research „Center for Scalable Data Analytics and Artificial Intelligence Dresden/Leipzig“, project identification number: ScaDS.AI 

We would also like to thank the German Federal Ministry for Digital and Transport for funding the DiGuRaL project as part of the mFund program.